%% file: root.tex
\documentclass[letterpaper, 10 pt, conference]{ieeeconf}  

\IEEEoverridecommandlockouts

\usepackage{microtype}				
\usepackage{cite}
\usepackage{amsmath,amsfonts} 
\usepackage[normalem]{ulem}
\usepackage{xpatch}
\usepackage{graphicx}
\usepackage{xcolor}
\usepackage{multirow}
\usepackage{colortbl}
\usepackage{makecell}
\usepackage{caption}
\usepackage{url}
\usepackage{multirow}
\usepackage{subcaption}
\usepackage{hyperref}

\title{\LARGE \bf
Serial Chain Hinge Support for Soft, Robust and Effective Grasp
}

\author{Dario Stuhne, Jelena Vuletić, Marsela Car and Matko Orsag
\thanks{Authors are with LARICS Laboratory for Robotics and Intelligent Control Systems, University of Zagreb, Faculty of Electrical Engineering and Computing, Unska 3, 10000 Zagreb, Croatia
        {\tt\small dario.stuhne, jelena.vuletic, marsela.polic, matko.orsag @fer.hr}}%
}

\begin{document}

\maketitle
\thispagestyle{empty}
\pagestyle{empty}

\begin{abstract}
This paper presents a serial chain hinge support, a rigid but flexible structure that improves the mechanical performance and robustness of soft-fingered grippers. Gravity can reduce the integrity of soft fingers in horizontal approach, resulting in lower maximum payload caused by a large deflection of fingers. To substantiate our claim we performed several experiments on payload and deflection of the SofIA gripper under both horizontal and vertical approach. In addition, we show that this reinforcement does not impede the original compliant behavior of the gripper, maintaining the original kinematic model functionality. We showcase the proprioceptive and exteroceptive capabilities (RGB-D camera and flex sensor) for two opposing manipulation problems: grasping small and large objects. Finally, we validated the improved SofIA gripper in agricultural and everyday activities.
\end{abstract}

\input{content/1_Introduction.tex}
\input{content/2-Hinge.tex}
\input{content/3-Perception.tex}
\input{content/4-Kinematics.tex}

\input{content/5_Conclusion.tex}


\bibliographystyle{ieeetr}
\bibliography{bibliography/bib}

\end{document}

%% file: content/1_Introduction.tex
\section{Introduction}
\label{sec:introduction}
\begin{figure}
\centering
\includegraphics[width=\linewidth]{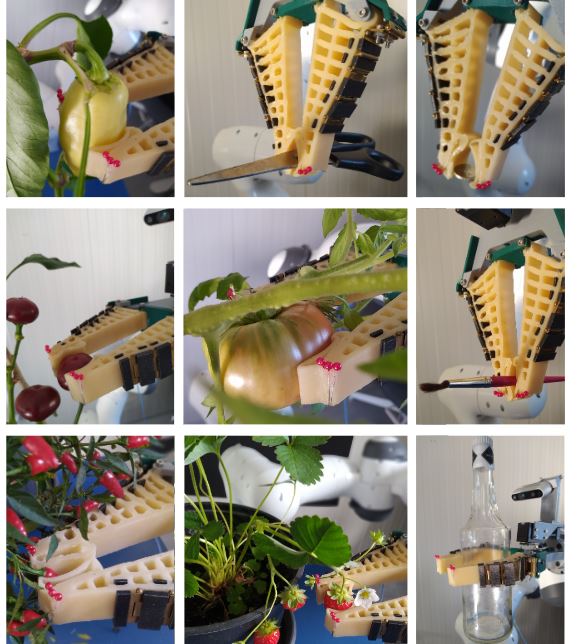}
\caption{SofIA soft gripper mounted on FRANKA EMIKA Panda robotic arm handling fruit (peppers, tomato, and strawberry) and miscellaneous everyday objects (a bottle, a coin, a paintbrush, scissors) in both horizontal and vertical approach. The serial chain hinge support can be seen on the outer side of the Fin-Ray finger structure.}
\label{fig:applications}
\vspace{-0.7cm}
\end{figure}

Driven by the ever-growing population and the accompanying food demand, impeded by climate change and labor shortage, traditional agriculture is rapidly changing \cite{Elfferich2022}. Observing the latest rise of robotics and automation in agriculture, it is safe to pin them as key drivers of this change. In spite of this latest rise of robotics in agriculture, it has yet to  reach the same level of success as in manufacturing industries. This is mostly because farming generally takes place in a very unstructured environment \cite{Zhang2020}, be it in open fields or in greenhouses. However, in contrast to open fields, greenhouses offer a certain level of organization and structure, which can expedite the use of robots in everyday tasks such as harvesting, picking, pruning, pollination, etc \cite{Polic2021}. Harvesting crops is a perfect example of challenges in modern agriculture \cite{Elfferich2022}, mainly because it is time-consuming and labor-intensive. Picked fruits and vegetables can vary in size, mass, and shape (Fig. \ref{fig:applications}), which requires dexterity and adaptability that can be overcome by implementing soft robotics solutions \cite{Tawk2019, Manti2015, Abondance2020}, which are being intensively studied in agriculture for harvesting applications \cite{Navas2021, Navas2021MDPI, Gunderman2022}. 

\begin{figure}
\centering
\includegraphics[width=\linewidth]{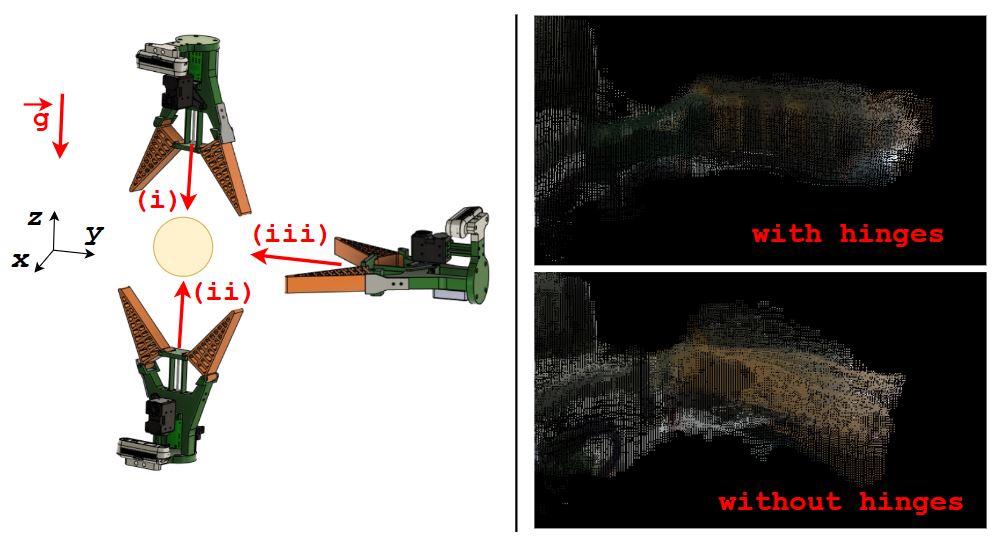}
\caption{\textbf{Left} - grasping approach directions with the respect to gravity. \textbf{Right} - the deflection in horizontal approach obtained from the point cloud for the gripper with and without the hinge support. The hinge support prevents larger deflection.}
\label{fig:approach-deflection}
\vspace{-1.1cm}
\end{figure}

The quality of harvesting and, in a more general case, crop handling depends on how the soft gripper performs against gravity. To compare the performance of soft grippers against gravity authors in \cite{Elfferich2022} group them into three categories (Fig. \ref{fig:approach-deflection}): (i) vertical approach from above, (ii) vertical approach from below, (iii) horizontal approach in the x-y plane (perpendicular to gravity). Only 15\% of the grippers studied showed the ability to operate in both horizontal and vertical approaches. This drawback is particularly pronounced in soft two-finger grippers, since a smaller number of fingers in the gripper, together with the softness of the material, can only lift smaller payloads in the horizontal direction. When comparing payloads of soft grippers for different approaches \cite{Zhou2018, Crowley2022},  researchers report a significant decrease in the maximum payload for the horizontal approach compared to the vertical approach. In \cite{Mathew2021}, 42\% less payload was reported when using a soft gripper for versatile and delicate gripping in the horizontal approach, which has a significant negative impact on various agricultural tasks in that direction. Clearly, to mitigate the undesirable effects of gravity on the horizontal approach, additional reinforcements/improvements to the grippers are required.

The challenge this paper addresses is to eliminate the negative effects of gravity in the horizontal approach without impeding the working capabilities of the gripper in other directions (Fig. \ref{fig:approach-deflection}). The research follows up on our previous work \cite{Stuhne2022} that presented a soft finger AI-enabled hand (SofIA, Fig. \ref{fig:applications}). Like other grippers reported in this paper SofIA's fingers deflect in a horizontal approach while  handling larger payloads \cite{Stuhne2022}. To solve this problem this study proposes using supporting hinges. As we show in the paper, this improves SofIA's capabilities in the horizontal approach, at the same time ensuring that when the gripper is reinforced, its compliance and softness have not deteriorated.

When it comes to handling delicate objects like picking ripe fruit, equally important as mechanical properties are the proprioceptive capabilities of soft grippers. Proprioception enables delicate closed-loop control during manipulation. In soft robotics it is often based on force and flex sensors \cite{wang2018, homberg2022}, but there have also been some efforts in the development of vision-based proprioceptive sensing \cite{polic2019, hundhausen2020}. SofIA encompasses both flex and optical sensors, which enable real-time fingertip position tracking, supplementing a proposed gripper kinematic model. A similar setup has been proposed in \cite{soter2018}, where flex sensor signals have been mapped to visual information from the camera, enabling posture tracking even when visual information is not available, e.g. due to occlusions. SofIA's sensory system is capable of the same sensory fusion. Additionally, the camera is positioned in a way that enables simultaneous finger posture and environment tracking, providing the possibility of exteroceptive sensing, which includes both grasp quality monitoring and target object detection and localization. 

Our main contribution is the novel serial chain hinge support structure that maximizes the payload and improves the mechanical properties of soft-fingered grippers. Additionally, we show how the supports can be used to increase the sensory apparatus of the SofIA gripper. To test this hypothesis we conducted various experiments with objects of different shapes and sizes, reporting the results supporting our hypothesis in Section \ref{sec:hinge}. In Section \ref{sec:perception} we demonstrate the mechanical properties and the perception capabilities in two opposing manipulation scenarios (i.e. manipulation of small and large objects). Putting it all together in Section \ref{sec:kinematics}, we show how hinges help improve the repeatability of the kinematic model of the gripper. We conclude the paper with examples of the variety of places in which SofIA has been benchmarked and validated such as RoboSoft 2022 competition and modern greenhouse.

%% file: content/2-Hinge.tex
\section{Serial Chain Hinge Support}
\label{sec:hinge}
Hinges have recently attracted research attention in the development of soft grippers. In \cite{Wang2017}, a shape memory alloy based gripper with two hinge segments on each finger made from Ni-Cr wire was designed and developed. The main reason for using hinges was the controllability via two hinges, resulting in a total of nine different finger configurations for gripping \cite{Wang2017}. Furthermore, a novel 3D-printed monolithic finger with built-in hinges for prosthetic applications was presented in \cite{Mutlu2016}. In this monolithic finger design, the hinges were attached to the finger in the form of indentations, representing the location where the finger bends when actuated. However, this approach suffers from the fatigue of the finger material, since the finger loses elasticity and quenches after certain opening/closing cycles. 

Compared to these existing attempts of integrating soft hinge-like structures, the study proposed in this paper is, to the best of our knowledge, the first attempt to utilize metal hinges as joints to reinforce the soft fingers. In this study hinges work as passive joints offering flexibility with their single degree of freedom (rotational joint between two plates). Ideally, hinges increase and improve the mechanical performance of SofIA's Fin-Ray fingers: the rigidity helps reduce the effects of deflection while spanning the range of maximum payload, with the initial softness and compliance of SofIA remaining intact due to the flexibility of joints connected in a serial chain.

\begin{figure}
\centering
\includegraphics[width=\linewidth]{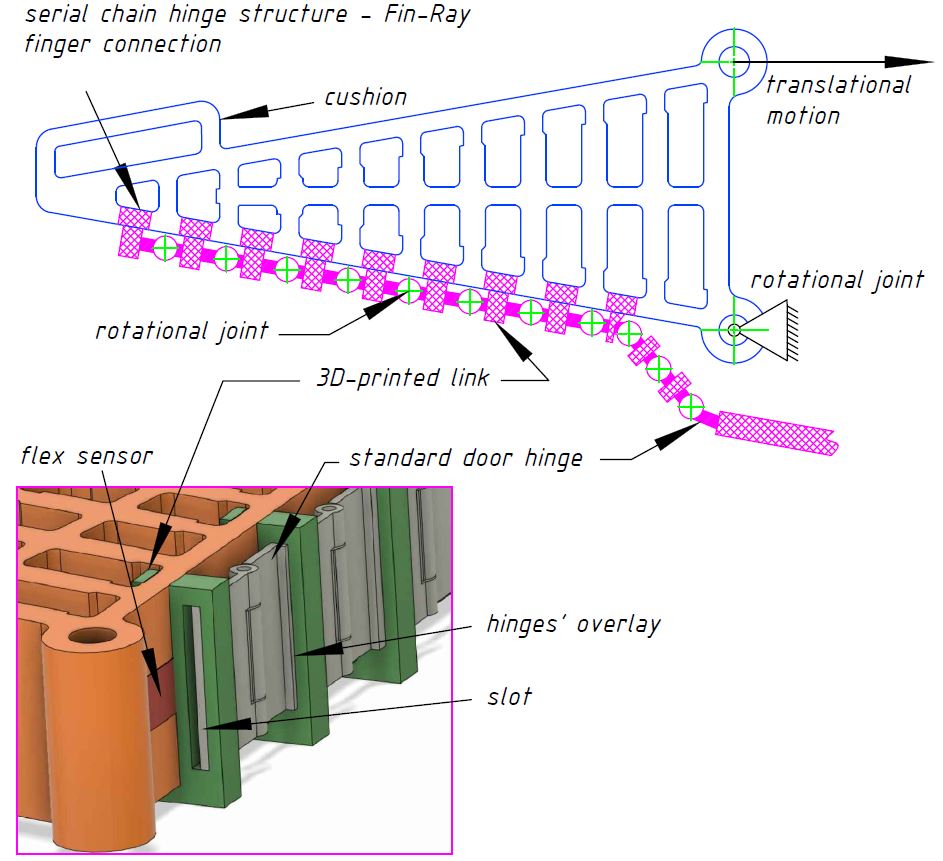}
\caption{Visualization of the implemented serial chain hinge structure on the side of the Fin-Ray finger. The serial chain hinge structure consists of 3D-printed links connecting standard door hinges with a rotational joint. The 3D-printed links are attached to the Fin-Ray finger with a clip-on mechanism. This way the serial chain hinge supports the gripper body and adds both rigidity and flexibility to the fingers to prevent deflection and maximize mechanical performance without sacrificing softness and compliance.}
\label{fig:hinge-concept}
\vspace{-0.7cm}
\end{figure}

This study builds upon our earlier work \cite{Stuhne2022} where SofIA was originally presented as a soft gripper designed mainly for agricultural purposes. To facilitate handling objects of larger sizes and masses, while maintaining the same general principles we increased the size and modified the original Fin-Ray finger \cite{Festo} with a longitudinal rib \cite{Stuhne2022}. Additional modification to the soft body structure from this study introduces a cushion on the fingertip to enable gripping of smaller objects (Fig. \ref{fig:hinge-concept}). The manufacturing procedure for the new fingers follows the same principles as presented in earlier our work.

SofIA is an underactuated soft gripper driven by a single DC motor, Dynamixel XM-430. Electric actuation facilitates control of the system with fewer components required compared to other actuation methods commonly used with soft grippers, i.e. pneumatic and hydraulic actuation methods.

To reduce deflection, the serial chain hinge structure must be attached to the body of the gripper. In this way, the forces and the bending moments that occur during grasping are taken over by the gripper's body, thus reducing the impact of payload and deflection on the fingers. For this purpose, the hinge chain is physically placed in the space between the lateral ribs on the outer side of the Fin-Ray finger structure, and the end of the chain is bolted to the side of the gripper body. The serial hinge chain consists of standard steel door hinges and custom 3D-printed links that serve both to join the hinges together and to attach the whole hinge chain support to the outer side of the Fin-Ray finger (Fig. \ref{fig:hinge-concept}). The hinge to hinge connection in these links is not rigid. It offers the neighboring hinges to slide one against the other, allowing a linear degree of freedom in every connection. This way the serial chain expands and contracts along with the soft Fin-Ray structure. In addition, a flex sensor, as a part of SofIA’s proprioceptive sensory system, can be installed between the hinge chain and the Fin-Ray finger to provide information about the finger bending angle.

The benefits of including hinges in SofIA are explored and validated in detail. For this purpose, SofIA is subjected to the payload and deflection study. Tests were performed with SofIA mounted on FRANKA EMIKA Panda robotic arm to obtain more credible results without human involvement where both SofIA and robotic arm were controlled via ROS interface.

\subsection{Payload}
\begin{figure}
\centering
\includegraphics[width=\linewidth]{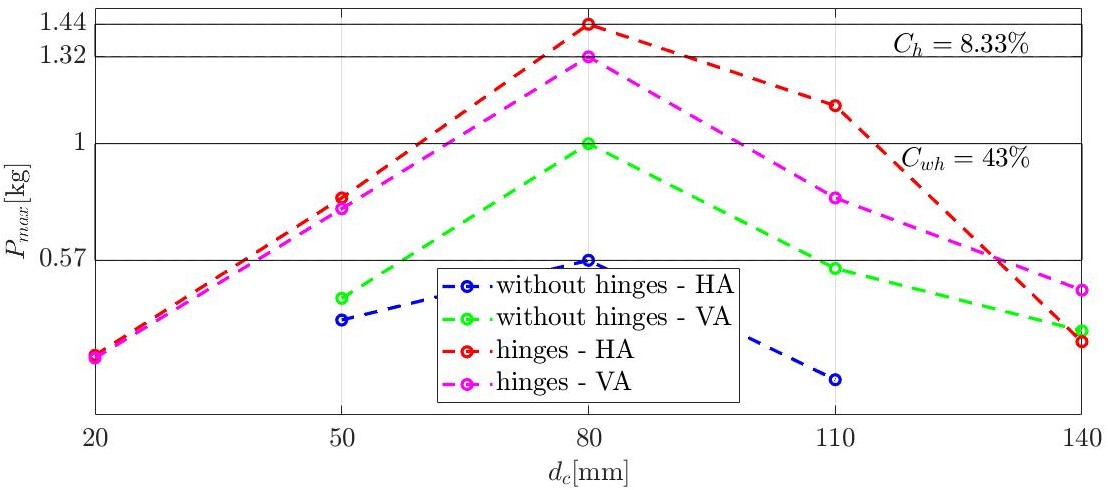}
\caption{Payload study for both configurations with and without hinges in horizontal approach (HA) and vertical approach (VA). This graph shows the relationship between the diameter of the grasped cylindrical object $d_{c}$ and the maximum payload $P_{max}$. Moreover, the differences between the payloads for each configuration can be observed in the graph with a relative change of only 8.33\% ($C_{h}=8.33\%$) for the maximum payloads in HA and VA for the finger configuration with hinges compared to the relative change of 43\% ($C_{wh}=43\%$) for the maximum payloads in HA and VA for the finger configuration without hinges.}
\label{fig:payload}
\vspace{-0.7cm}
\end{figure}

An extensive study on the effect of serial chain hinge support on SofIA payload is conducted in a series of experiments. The first experiment investigates the gripper payload capabilities depending on the grasped object size. A set of cylindrical primitives with varying diameter and weight were used as payload objects in 4 experimental setups, namely for a gripper with and without hinge support, and in both horizontal and vertical approach. Payload for each experimental setup was obtained as the average of four measurements for each object using a hand scale with an accuracy of 0.01 kg suspended on the object. The limitation of this method is that the hand scale has its own mass (0.12 kg), which is more than the maximum payload for a horizontal approach with an 140 mm in diameter object (Fig. \ref{fig:payload}). The maximum payload was determined visually on the hand scale. It was also not possible to measure the payload for a finger configuration without hinges for a small 20 mm in diameter object because the fingers tend to twist. The increase in the length of the finger reported in \cite{Stuhne2022} decreased the torsional stiffness of the Fin-Ray finger, making it more prone to twist.

Fig. \ref{fig:payload} shows the identified relationship between the cylindrical objects diameter $d_{c}$ and the maximum payload $P_{max}$. As expected the optimal performance is achieved for the medium-sized objects in accordance with other similar studies \cite{Sun2022}. Deteriorated performance for smaller objects can be explained by lower contact forces between the object and the finger due to the smaller deformation of the finger. Similarly, the performance deteriorates for larger objects when the actuator reaches its limits. The findings more relevant for this study are related to the comparison of maximum payloads for cases with and without hinges. Considering the original SofIA design without hinges, we observe a higher maximum payload for the vertical approach compared to the horizontal approach (43\% higher). The deflection of the fingers increases together with the object's mass and size, which negatively affects mechanical performance during the horizontal approach. The introduction of hinges improves the overall payload capabilities, increasing the maximum payload in both approach directions. An interesting finding of this study is the higher maximum payload of the horizontal approach (8.33\%) for the SofIA gripper with hinges.

\subsection{Deflection}
\begin{figure}
\centering
\includegraphics[width=\linewidth]{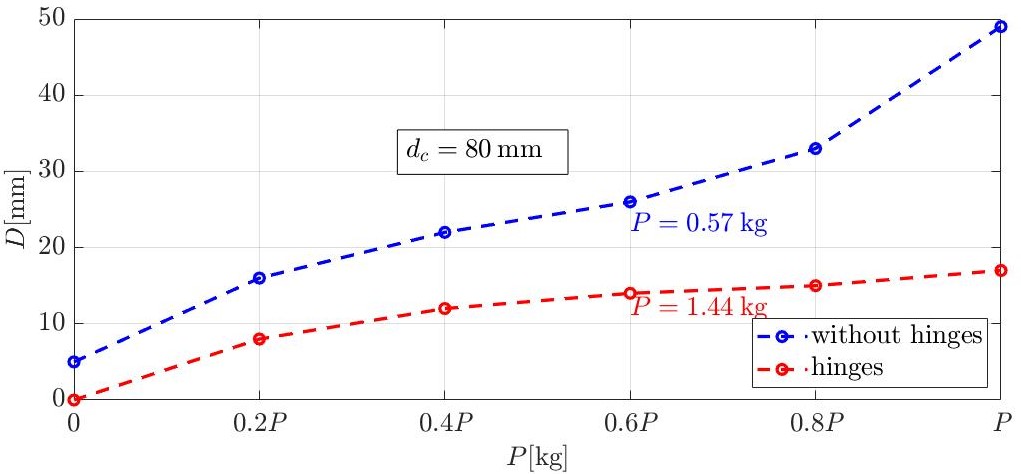}
\caption{Finger deflection study for both configurations with and without hinges under maximum payload $P$ for both respective configurations in Fig. \ref{fig:payload}. This graph represents the amount of deflection $D$ with respect to the incremental change in the payload (0, $0.2P$, ...,$P$) for the use of the SofIA in the horizontal approach (HA). The deflection increases when increasing payload for both configurations, but the overall deflection is significantly higher for the configuration without hinges.}
\label{fig:deflection}
\vspace{-0.6cm}
\end{figure}

The deflection, which is observed only in the horizontal approach, is not completely eliminated by adding hinges to the side of the finger, but it is visibly smaller and less pronounced than without the hinge support. Fig. \ref{fig:deflection} shows the relationship between the payload and the deflection for each finger configuration. Previous study showed that both configurations exhibit optimal payload capabilities for 80 mm in diameter object (Fig. \ref{fig:payload}). Therefore, the deflection experiment was conducted by gradually increasing the weight of the 80 mm in diameter object. The results in Fig. \ref{fig:deflection} clearly show that the deflection is significantly smaller for the finger configuration with hinges. These results allow us to conclude that hinges significantly improve the mechanical properties of the gripper.

%% file: content/3-Perception.tex
\section{Perception}
\label{sec:perception}
SofIA's sensory system consists of an RGB-D camera, Intel RealSense D435, and a Spectra Symbol flex sensor. The RGB-D camera is mounted so that both exteroception (environment sensing) and proprioception (grasp quality assessment and kinematic behaviour monitoring) are available simultaneously. The proposed serial chain hinge structure enabled inserting a flex sensor directly between the finger and the hinge structure, thus expanding the sensory apparatus of the gripper and enabling proprioceptive capabilities of SofIA. Proprioception has been exploited both for validation of the kinematic model, and as a supplement for the kinematic model, which is further discussed in Section \ref{sec:kinematics}. Flex sensor data can be used to track the finger posture even when the grasped object occludes the fingertips and impairs camera-based proprioception. On the other hand, SofIA's exteroceptive capabilities are crucial for planning, as executing a stable grasp of objects of different sizes requires estimation of the size of an object which is to be grasped. Additionally, SofIA's choice between the two possible approach directions (horizontal and vertical) depends on the target object's dimensions.

In our previous work \cite{Polic2021}, we used a convolutional neural network to detect and harvest peppers by extracting grasping target dimensions from the depth image recorded with an RGB-D camera. This approach worked well for agricultural applications and fruit picking. The same approach was simplified for the RoboSoft 2022 competition which was another opportunity to benchmark the gripper. In all of the tasks, various objects that had to be detected autonomously were placed on a flat surface in front of the robotic manipulator. First, we recorded point clouds from multiple angles, transformed them from the camera frame to the global coordinate frame, and registered the transformed point clouds. The registered point cloud was then filtered based on a priori known information about the environment, i.e. the bounding boxes in which objects could be placed. The remaining filtered points were used to estimate the object dimensions and to decide on the grasping approach based on the most dominant object dimension, taking into consideration the kinematic limitations of the robotic manipulator.

%% file: content/4-Kinematics.tex
\section{Kinematic Model}
\label{sec:kinematics}
\begin{figure}
\centering
\includegraphics[width=\linewidth]{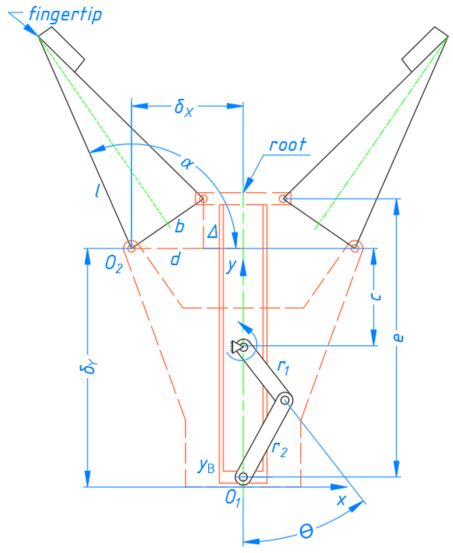}
\caption{Geometric relationships between the main features of SofIA used to derive the kinematic model. During the actuation of the motor, the position of the slider $\Delta$ changes, resulting in the rotation of the fingertip around the point $O_{2}$. Knowing the exact mathematical expression of the change in the fingertip position with respect to the change in DC motor rotation angle $\theta$ enables precise grasping of both larger and smaller objects.}
\label{fig:kinematics}
\vspace{-0.6cm}
\end{figure}

Slider-crank linkage, driven by a DC servomotor, is a core mechanism that enables finger movement in the SofIA gripper. It converts rotational motion into translational, moving the position of $y_B$ along the $y$-axis (Fig. \ref{fig:kinematics}):
\begin{equation}
    y_B = r_1 \cdot \textup{cos}(\theta)+\sqrt{r_2^2 - r_1^2 \cdot \textup{sin}^2(\theta)},
\end{equation}
where $\theta$ stands for the input motor angle and $r_1$ and $r_2$ denote the length of the crank and the rod connecting the crank and the slider, respectively. Controlling the position of $y_B$ moves the slider position $\Delta$ and shapes the finger base length $b$:
\begin{gather}
    \label{eq:delta}
    \Delta = e-c-y_B \\
    b = \sqrt{d^2 + \Delta^2}.
\end{gather}
The fingers are modeled as isosceles triangles with constant leg length $l$ and variable base angle and base length $b$. The change in the base angle of the soft finger is taken into account when calculating the fingertip angle of rotation $\alpha$ around the point $O_2$:
\begin{equation}
    \alpha = \textup{arcsin}(\frac{\Delta}{b}) + \textup{arccos}(\frac{b}{2l}).
\end{equation}
Finally, the positions of both the left and the right finger are calculated as follows:
\begin{gather}
        x_{left} = l \cdot \textup{cos}(\alpha) - \delta_x \\
        x_{right} = \delta_x - l \cdot \textup{cos}(\alpha) \\
        y_{left} = y_{right} = l \cdot \textup{sin}(\alpha) + \delta_y,
\end{gather}
where $\delta_x$ and $\delta_y$ denote distances from $O_2$ to the center of SofIA coordinate system $O_1$.

\begin{figure}
\centering
\includegraphics[width=\linewidth]{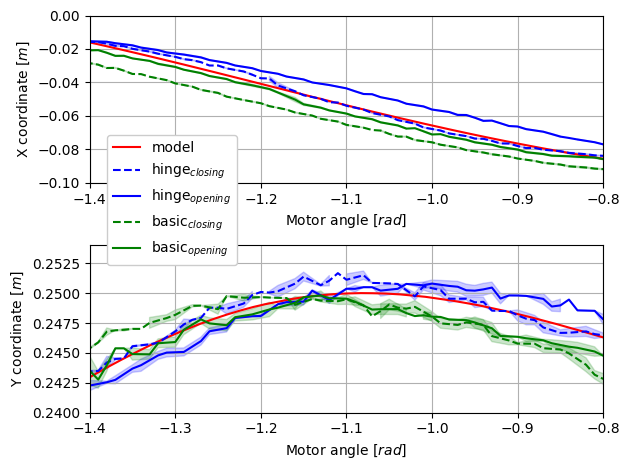}
\caption{SofIA kinematics shown for the left finger with and without the serial chain hinge structure, compared to the proposed kinematic model. The position of the fingertip was recorded in 10 experiment repetitions for both structures, where a single repetition consisted of both opening and closing the hand.}
\label{fig:kinematika_graf}
\vspace{-0.7cm}
\end{figure}

In order to compare the ideal model with the actual behavior of both the hinge-reinforced and the original version of SofIA, fingertip position was recorded using SofIA's proprioceptive capabilities. Fingertips were marked with red pins and thus straightforwardly filtered in HSV color space in the image recorded with Intel RealSense D435 camera. SofIA was repeatedly opened and closed in increments of 0.015\,rad. Position of the fingertip was extracted in the image for each increment, and its 3D position in the camera coordinate frame was obtained using the corresponding registered point cloud. Finally, as the camera was calibrated using the procedure described in \cite{maric2020unsupervised}, the transformation from the SofIA coordinate frame to the camera coordinate frame ${\textbf{T}_S^C}$ was known a priori and used to transform the fingertip position from the camera coordinate frame to the SofIA coordinate frame. Fingertip positions in both opening and closing sequences, along with the dissipation across 10 experiment repetitions, for both SofIA configurations are shown in Fig. \ref{fig:kinematika_graf}.  Maximum and minimum motor angles on Fig. \ref{fig:kinematika_graf} correspond to the fully open (-0.8\,rad) and the fully closed (-1.4\,rad) gripper positions, respectively.

As can be seen in Fig. \ref{fig:kinematika_graf}, SofIA gripper reinforced with the proposed hinge structure follows the derived kinematic model more precisely than the original not reinforced version. While the deviation from the model on Y coordinate is on a millimeter scale for both structures, differences along X axis reach up to 1\,cm in opposite directions for different structures. On one hand, the hinges tend to prevent the gripper from reaching the fully open position, resulting in lower X coordinate values for the left fingertip position, while on the other hand, the finger without the supporting structure exceeds the modeled maximum gripper opening. The deviation between the closing and the opening sequence is also observed in both cases and is a consequence of backlash in mechanism. 

\begin{figure}
\centering
\includegraphics[width=\linewidth]{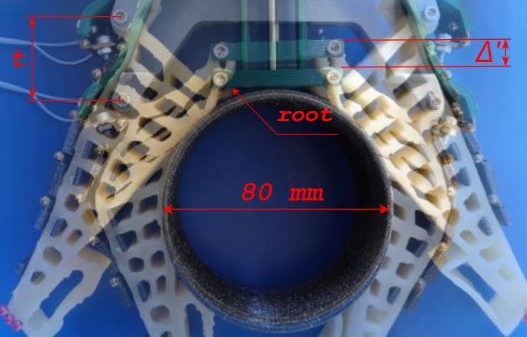}
\caption{Overlay of fully open gripper before the grasp and fully closed gripper grasping the target object. Robotic manipulator translated along the grasping axis for $t$ in order to compensate for total slider displacement $\Delta$. Slider displacement $\Delta'$ remained uncompensated.}
\label{fig:grasping}
\vspace{-0.3cm}
\end{figure}

Considering that both the finger length (Y coordinate in Fig. \ref{fig:kinematika_graf}, i.e. the tip of the grasp) and the slider position $\Delta$ (Eq. \ref{eq:delta}, i.e. the root of the grasp) vary as a function of the joint, grasping objects of different sizes requires different approaches. Two edge cases, namely grasping of large ($\ge$ 80 mm in diameter) and small ($\le$ 10 mm height) objects, have been examined as a part of this study, demonstrating improvement introduced with the serial chain hinge support, while retaining compliant properties of the original version of SofIA. A stable grasp of bigger objects has to be enveloping, i.e. the entire volume of the target object has be contained between the fingers. In order to achieve that, the motion of the root of the grasp $\Delta$ has to be compensated with the robotic arm motion $t$. The SofIA's mounting point must move in the opposite direction along the same axis while closing the gripper (Fig. \ref{fig:grasping}). That way, the grasped object will be positioned as close to the gripper body as possible, ensuring maximum grasp stability. For small-size objects, a successful grasp can be achieved thanks to the cushions attached to the fingertips (Fig. \ref{fig:hinge-concept}). However, when grasping small objects, it has to be ensured that the grasped object is positioned between the centers of the cushions once the gripper is fully closed. As the position of the fingertips on $y$-axis decreases when closing the gripper, the robotic arm has to compensate for the displacement of the fingertips, which can easily be achieved if the proposed kinematic model is taken into account when planning the grasp.

\begin{figure}
\centering
\includegraphics[width=\linewidth]{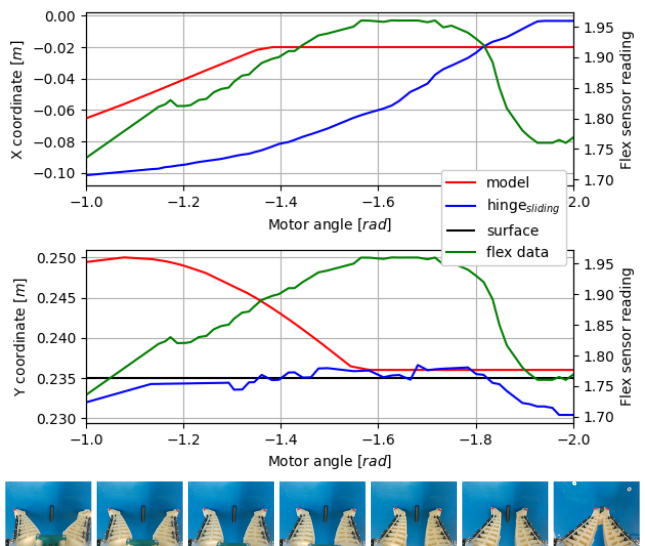}
\caption{Position of the tip of the left finger for the reinforced SofIA in a sliding mode, optimal for small object grasping, visualized along with flex sensor readings, position obtained from the kinematic model and position of a flat surface on which an object is placed. Finger positions recorded during the grasping procedure are shown under the graph.}
\label{fig:sliding}
\vspace{-0.7cm}
\end{figure}

A special case of small object grasping occurs when objects small in height are placed on a flat surface, which was one of the tasks in RoboSoft 2022 competition and in many other common manipulation tasks such as small parts assembly. Such objects have to be approached in vertical direction in order to avoid collision of the manipulator with the flat surface during manipulation. In the proposed scenario, in order for grasp to be successful, the fingertip position in $y$-axis in SofIA coordinate frame calculated using the proposed model would be larger than the $y$-axis coordinate of the flat surface $y_S$, detected in the same coordinate frame. Passive compliance of SofIA fingers enabled successful execution of small object grasping task by sliding along the surface while approaching the grasping pose. Position of the fingertip in sliding mode is shown in Fig. \ref{fig:sliding}. It can be seen that the gripper does not fully close at -1.4\,rad, but it instead reaches fully closed position at -1.9\,rad. At this point, the object is successfully grasped and the upper part of the finger is no longer bent. During the whole grasping procedure, the fingertip position in $y$-axis is approximately constant and corresponds to $y_S$. 

Fig. \ref{fig:sliding} also shows flex sensor readings recorded during the sliding grasp. It can be seen that the bending angle increases during the approach and settles once the sliding is terminated, validating the use of flex sensor for proprioceptive real-time sensing. In the described sliding mode procedure, kinematic model provides information on where the robotic manipulator should be positioned in order for grasp to be successful, i.e. in which manipulator pose will the fingers, once closed, successfully grasp the target object. Meanwhile, flex sensor enables feedback control during delicate small object grasping, providing information in which direction the manipulator should move. In specific gripper working regimes, such as sliding mode, where fingers do not follow the kinematic model, supplementing the model with proprioceptive sensory system measurements is of great importance for achieving precise delicate feedback control.


%% file: content/5_Conclusion.tex
\section{Conclusion}
\label{sec:conclusion}
SofIA, a soft gripper inherently designed for agricultural applications, was validated on agricultural tasks such as picking peppers, strawberries, and tomatoes. Other agrotechnical use cases we tested SofIA on was picking and pollinating flowers for yield control. SofIA proved to be capable of very delicate operations, like holding flowers for pollination and picking strawberries. It also performed well when picking tomatoes, which were significantly heavier than strawberries, and required the support of the hinges. To formally benchmark the capabilities of the serial chain hinge structure, we went a step further to test SofIA in three different scenarios defined as a part of the 2022 RoboSoft competition: (i) pick and place task, (ii) potted plant transportation, (iii) pouring from a bottle and serving. All tasks are general in nature and involve everyday activities, and the final result was intended to show SofIA's robustness and effectiveness. We have successfully performed all the tasks (for more details please see the supplemental video and the \href{https://www.youtube.com/playlist?list=PLC0C6uwoEQ8YZfoCRMe0X0HLZsRzIMQCY}{Youtube video \cite{LaricsLab}}). We attribute this to the serial chain hinge structure that maximized the mechanical performance and facilitated exteroception-based algorithms that we used to plan all the grasps and approaches autonomously.

The experiments presented in this study show that adding the serial chain hinge support improved the mechanical performance and the robustness of the SofIA gripper. The hinge structure enhanced SofIA's capabilities in both the horizontal and vertical approach in terms of: (i) achieving larger payloads (32\% more in the vertical approach and 252\% more in the horizontal approach), (ii) reducing the relative changes between payloads in horizontal and vertical approaches (43\% relative change for a finger configuration without support and only 8.33\% for a finger configuration with hinge support), and (iii) reducing the gradient of deflection in horizontal approach. The proposed hinge structure does not impede the kinematics model of the original soft gripper, with minor deviations compared to the ideal mathematical model. Finally, we showed how this mechanical improvement, the kinematics model, and the additional perceptive capabilities of the gripper enable it to perform well in two opposing scenarios: grasping small and large objects.